# BAYESLDM: A DOMAIN-SPECIFIC MODELING LANGUAGE FOR PROBABILISTIC MODELING OF LONGITUDINAL DATA

**Karine Tung** [1]  **Steven De La Torre** [2]  **Mohamed El Mistiri** [3]  **Rebecca Braga De Braganca** [2]  **Eric Hekler** [4]
**Misha Pavel** [5]  **Daniel Rivera** [3]  **Pedja Klasnja** [6]  **Donna Spruijt-Metz** [2]  **Benjamin M. Marlin** [1]

## ABSTRACT

In this paper we present BayesLDM, a system for Bayesian longitudinal data modeling consisting of a high-level modeling language with specific features for modeling complex multivariate time series data coupled with a compiler that can produce optimized probabilistic program code for performing inference in the specified model. BayesLDM supports modeling of Bayesian network models with a specific focus on the efficient, declarative specification of dynamic Bayesian Networks (DBNs). The BayesLDM compiler combines a model specification with inspection of available data and outputs code for performing Bayesian inference for unknown model parameters while simultaneously handling missing data. These capabilities have the potential to significantly accelerate iterative modeling workflows in domains that involve the analysis of complex longitudinal data by abstracting away the process of producing computationally efficient probabilistic inference code. We describe the BayesLDM system components, evaluate the efficiency of representation and inference optimizations and provide an illustrative example of the application of the system to analyzing heterogeneous and partially observed mobile health data.

*Keywords*— Bayesian inference, probabilistic programming, time series, missing data, Bayesian imputation, mobile health

## 1 INTRODUCTION

Effective modeling of complex multivariate temporal processes is a key methodological challenge in machine learning and statistics. This paper is particularly motivated by challenges that arise when modeling longitudinal data under extensive missingness, high heterogeneity, and general data scarcity. Different combinations of these challenges occur in many time series modeling application domains including astronomy [21], economics [8], finance [12], and meteorology [15], but these challenges often all occur simultaneously in the health data analysis setting [13].

Indeed, mobile health [9] is a key motivating domain for this work as data sets from this domain often include elements from multiple heterogeneous sources with very different properties (e.g., wearable sensor measurements and behavioral self-report [22]), are subject to extensive missingness (e.g., due to self-report non-response, sensor non-wear, etc. [6]), and are often relatively low-volume due to the high cost and complexity of running field studies for long time periods and/or with many participants.

In this paper, we respond to these key longitudinal data modeling challenges by presenting the Bayesian Longitudinal Data Modeling (BayesLDM) system. BayesLDM consisting of a high-level probabilistic modeling language with specific features for modeling complex multivariate time series data coupled with a compiler that produces a probabilistic program for performing inference in the specified model. The BayesLDM modeling language helps to close the gap between what a modeler would write on paper or sketch on a whiteboard and the formal specification of a model. We provide an overview of the BayesLDM system in Figure 1.

The model family that BayesLDM supports is Bayesian networks [18], which subsume many classical models including linear regression, logistic regression, factor analysis and mixture models. BayesLDM has a specific focus on the efficient declarative specification and implementation of dynamic Bayesian Networks (DBNs) [16]. DBNs are probabilistic dynamic models that generalize vector autoregression [11] and a variety of models classically used in the social sciences including path diagrams [25].

The BayesLDM compiler combines the model specification with data inspection and outputs code for performing Bayesian inference for unknown model parameters and missing data using an underlying probabilistic inference library [5]. The focus on Bayesian inference methods has a number of well-known advantages including exposing posterior uncertainty over unknown model parameters due to low volumes of data (particularly when fitting individual-level models) [5]. BayesLDM also uses the same inference methods to automatically handle missing data using Bayesian imputation.

BayesLDM currently supports compiling BayesLDM models into Python programs using NumPyro as the probabilistic inference library [19]. Relative to NumPyro, the BayesLDM modeling language provides significantly higher-level modeling abstractions including language features for efficiently specifying dynamic model structures. Further, the BayesLDM compiler has the ability to produce optimized inference implementations for DBNs and

---

[1]University of Massachusetts Amherst, Amherst, MA, USA  [2]University of Southern California, Los Angeles, CA, USA  [3]Arizona State University, Tempe, AZ, USA  [4]University of California San Diego, San Diego, CA, USA  [5]Northeastern University, Boston, MA, USA  [6]University of Michigan, Ann Arbor, MI, USA

BayesLDM: A Domain-Specific Language for Probabilistic Modeling of Longitudinal Data

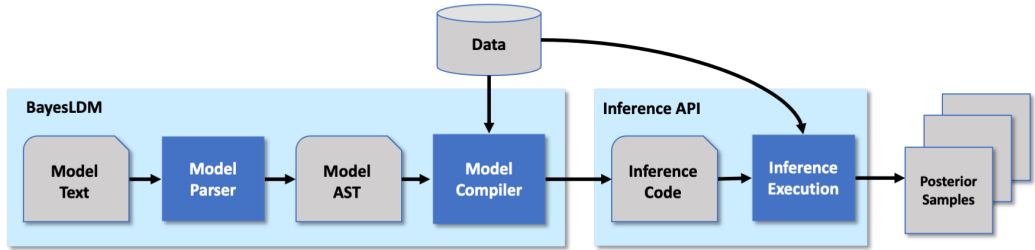

*Figure 1.* Overview of BayesLDM: Models are specified in the BayesLDM modeling language. The BayesLDM parser converts the model text into an abstract syntax tree (AST) representation. The BayesLDM model compiler uses the AST representation of the model and the provided data to produce inference code, including automated marginalization of incomplete data. Inference code could be output for execution using any underlying Bayesian inference library. In this work, we output inference code for execution using the NumPyro library. Importantly, the BayesLDM compiler has the ability to provide inference optimizations for certain model structures within general Bayesian network models and produces highly optimized inference implementations for Dynamic Bayesian Network models by leveraging advanced NumPyro language features.

also provides inference optimizations for certain model structures within general Bayesian network models (see Section 4). By leveraging NumPyro's more efficient but more complex features, these optimizations allow BayesLDM to achieve significantly faster inference run times compared to implementations of DBN model structures implemented directly in NumPyro using its more basic and user-friendly features.

In summary, BayesLDM's data inspection and model compilation capabilities automate the lower-level code optimization tasks required to achieve performant inference implementations when constructing time series models and handling missing data. In particular, the BayesLDM system more completely separates the specification of a model from the implementation and execution of inference computations resulting in model specification code that is both significantly more compact and more easily modifiable. As a result, BayesLDM has the potential to significantly accelerate the model development and evaluation cycle by reducing the time needed to produce correct and efficient inference implementations for longitudinal models and complex patterns of missing data.

The primary contributions of this work are thus: (1) The design of an intuitive and expressive probabilistic modeling language with specific language features for modeling complex longitudinal data. (2) The development of an optimizing compiler for this language that can convert models into efficient inference code for execution using a probabilistic inference library while automating marginalization of missing data during inference. (3) The conduct of experiments examining the representational and implementation efficiency of BayesLDM relative to related probabilistic programming tools. (4) The presentation of a case study applying the system to the analysis of heterogeneous and partially observed mobile health data.

The remainder of this paper is organized as follows: Section 2 presents background and related work. Section 3 describes the BayesLDM modeling language. Section 4 describes the BayesLDM compiler. Sections 5 to 7 present evaluations of the BayesLDM modeling language and optimizing compiler including the mobile health case study. BayesLDM is available for download at https://github.com/reml-lab/BayesLDM.

## 2 BACKGROUND

In this section we provide background material on time series data, handling missing data, Bayesian network models and Dynamic Bayesian Network models, Bayesian inference methods and probabilistic programming languages.

### 2.1 Multivariate Time Series

In a general multivariate data set $\mathcal{D} = \{\mathbf{x}_n \mid 1 \leq n \leq N\}$, each data case $\mathbf{x}_n$ corresponds to a $D$-dimensional vector $\mathbf{x}_n = [x_{1n}, ..., x_{Dn}]$. In the domains of primary interest in this work, a data set $\mathcal{D} = \{\mathbf{x}_n \mid 1 \leq n \leq N\}$ contains a collection of $N$ multivariate time series $\mathbf{x}_n$. We assume that each multivariate time series $\mathbf{x}_n = [x_{1n}, ..., x_{Dn}]$ consists of $D$ univariate time series $x_{dn}$. We assume the discrete time setting where the time points at which data values can be observed are regularly spaced and shared by all time series in the data set.

Each univariate time series $x_{dn} = [x_{dn}[0], ..., x_{dn}[L_n - 1]]$ is assumed to be of length $L_n$ with $x_{dn}[t]$ being the value at time $t$. For example, in a health domain application, $x_{1n}$ might correspond to the time series of heart rate values for patient $n$ while $x_{2n}$ might correspond to the time series of blood pressure values. We note that when discussing examples, it will often be more useful to provide more informative names for the variables that each dimension represents such as $h_n$ and $b_n$ for time series of heart rate and blood pressure values for patient $n$.

### 2.2 Missing Data

When implementing models, we use a data frame abstraction that supports the use of floating point NaN values to indicate when a specific time series value $x_{dn}[t]$ is missing. Figure 2 illustrates a multivariate time series $\mathbf{x} = [y, z]$ consisting of incomplete time series for each of two variables $y$ and $z$. In this visualization, missing values are represented by breaks in the line plot. Mathematically, we use the notation $\mathbf{x}_n^o$ to refer to all observed values in the multivariate time series $\mathbf{x}_n$ and $\mathbf{x}_n^m$ to refer to all missing values [10]. Similarly, we define $\mathcal{D}^o = \{\mathbf{x}_n^o \mid 1 \leq n \leq N\}$ and $\mathcal{D}^m = \{\mathbf{x}_n^m \mid 1 \leq n \leq N\}$ to represent all observed data in the data set and all missing data in the data set.

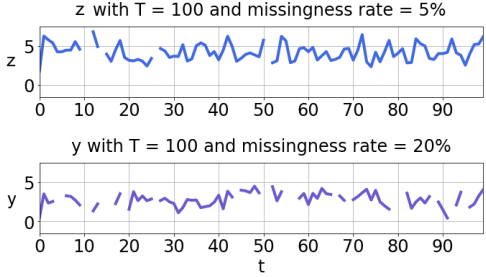

Figure 2. Example of a multivariate time series consisting of incomplete observations of two variables $y$ and $z$.

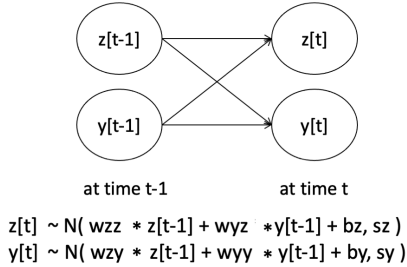

z[t] ~ N( wzz * z[t-1] + wyz * y[t-1] + bz, sz )
y[t] ~ N( wzy * z[t-1] + wyy * y[t-1] + by, sy )

Figure 3. Example of a basic DBN model.

## 2.3 Bayesian Networks

Bayesian Networks [18] provide a factorized parametric joint distribution $p(\mathbf{X} = \mathbf{x}|\theta)$ over a $D$-dimensional vector-valued random variable $\mathbf{X} = [X_1, ..., X_D]$ and are used to model general multivariate data. Bayesian networks encode the probabilistic independence structure of this joint distribution using a directed acyclic graph (DAG) $G$. The nodes in $G$ are the individual random variables $X_d$ and the directed edges indicate the presence of direct probabilistic dependencies. We define the function $\text{Pa}_G(X_d)$ to return the parents of random variable $X_d$ in graph $G$. The factorization of the joint distribution is then given in Equation 1 where $\theta$ are the model parameters. Bayesian network models are very flexible. They can model both continuous and discrete random variables and the distributions used to specify individual factors can be of any form.

$$p(\mathbf{x}|\theta) = \prod_{d=1}^{D} p(x_d|\text{Pa}_G(x_d), \theta) \quad (1)$$

## 2.4 Dynamic Bayesian Networks

Dynamic Bayesian networks (DBNs) extend Bayesian networks to the case where there is a time series of observations for each variable [16]. They are used to model multivariate time series data. Specifically, the DAG $G$ for a DBN includes edges both within time slices $t$ and between pairs of time slices $t$ and $t + \delta$, $\delta \geq 1$. DBN models with regular structure through time can be represented as a graph defined over a consecutive set of time slices starting from an initial time point $t$. Figure 3 specifies a DBN graph for the data introduced in Figure 2. This model corresponds to a vector autoregressive structure on the two variables $\mathbf{y}$ and $\mathbf{z}$ that comprise the multivariate time series. In the model, both variables directly influence each other at the next time step only, so it suffices to visualize two time slices of the graph.

## 2.5 Bayesian Inference

Given a Bayesian network or DBN model $p(\mathbf{x}|\theta)$ and a corresponding data set $\mathcal{D}$ as described in the previous sections, our goal is to infer the parameters of the model. For simplicity in this section, we will assume that all data values are real numbers, but all development in this section can be adapted to the case of discrete-valued data variables. The traditional approach to performing parameter inference in such models is maximum likelihood estimation (MLE), as described in the following optimization problem.

$$\hat{\theta}_{MLE} = \underset{\theta \in \Theta}{\text{argmax}} \sum_{n=1}^{N} \log\ p(\mathbf{x}_n|\theta) \quad (2)$$

This MLE problem is usually computationally efficient for Bayesian network and DBN models when the individual factor distributions belong to standard families and all of the data variables are fully observed. However, in the presence of missing data, the maximum likelihood principle combined with the missing at random assumption requires selecting the parameters that make the *observed* data the most likely [10]. This requires marginalizing over the missing data, which can result in significant additional computational complexity during learning as well as the need to approximate potentially intractable integrals. Further issues such as overfitting can arise in the MLE approach when data are scarce.

For these reasons, BayesLDM is based on approximate Bayesian inference methods. Approximate Bayesian inference treats unknown model parameters as random variables and approximates the posterior distribution over the model parameters given the observed data $p(\theta|\mathcal{D}^o)$. This posterior distribution directly captures the uncertainty in the unknown parameter values resulting from both missingness and data scarcity. Bayesian inference is based on the application of Bayes' theorem, as shown below.

$$p(\theta|\mathcal{D}^o) = \frac{p(\theta, \mathcal{D}^o)}{p(\mathcal{D}^o)} = \frac{p(\mathcal{D}^o|\theta)p(\theta)}{p(\mathcal{D}^o)} \quad (3)$$

First, note that unlike in the case of maximum likelihood estimation, Bayesian inference requires the specification of a distribution $p(\theta)$ on the model parameters $\theta$, referred to as the prior distribution. This distribution encodes any prior beliefs about the values of the model parameters and is an opportunity to encode additional prior knowledge about the problem domain.

For most models the exact posterior $p(\theta|\mathcal{D}^o)$ is computationally intractable, resulting in the need for approximation methods. BayesLDM leverages Markov chain Monte Carlo (MCMC) methods that are able to efficiently sample from the model posterior $p(\mathcal{D}^o|\theta)$. In particular, Hamiltonian Monte Carlo (HMC) makes use of gradients of the function $p(\mathcal{D}^o|\theta)p(\theta)$ to guide the sampling procedure and improve sampling efficiency [17]. The No-U-Turn Sampler (NUTS) automatically adapts tuning parameters in the HMC algorithm [7] and is used in several probabilistic programming languages including Stan [2], Pyro [1] and NumPyro [19]. NUTS is the default MCMC method used in BayesLDM.

Importantly, we note that in the presence of missing data, we can perform MCMC-based inference for the unknown model parameters and the missing data values simultaneously using MCMC methods, effectively approximating the joint posterior distribution $p(\theta, \mathcal{D}^m|\mathcal{D}^o)$. By discarding posterior samples for $\mathcal{D}^m$, we are left with samples of the parameter posterior only. The required MCMC computations are tractable if the function $p(\mathcal{D}^o, \mathcal{D}^m|\theta)$ can be efficiently computed. This is the case for Bayesian network and DBN models as this function specifies values for all data

variables, eliminating the need to perform explicit marginalization. Finally, as a byproduct, this approach also produces multiple samples for the missing data values $\mathcal{D}^m$, which correspond to multiple imputations under the model (again, under the assumption that missing data are missing at random). We note that while the NUTS algorithm can only be applied to continuous random variables, BayesLDM supports inference for missing discrete random variables using enumeration approaches provided by the NumPyro library.

## 2.6 Probabilistic Programming Languages

An important goal of probabilistic programming languages (PPLs) is to separate the specification of a probabilistic model from the code required to perform inference, learning and prediction using that model. Probabilistic programming languages use a variety of universal inference methods to enable this including MCMC and variational inference methods [24]. There is a wide array of general purpose PPL tools available currently including Pyro [1], NumPyro [19], Stan [2], PyMC3 [20], Edward2 [23], and Turing [4].

By contrast, BayesLDM is a modeling language that focuses specifically on Bayesian networks with specific language features for enabling the efficient specification of Dynamic Bayesian Networks. Importantly, BayesLDM seeks to directly leverage the strengths of existing work on general purpose probabilistic programming languages by optimizing and compiling BayesLDM model specifications into inference code that can be interpreted or executed by existing efficient probabilistic inference libraries, including PPLs. The current version of BayesLDM compiles models into Python code using NumPyro as the general purpose inference library.

# 3 THE BAYESLDM LANGUAGE

The BayesLDM language is formally specified using a grammar defined using a meta language tool specifically targeted at the development of domain specific languages. In this section, we describe some of the key features of the BayesLDM language.

## 3.1 Constants, Variables, Operators, etc.

BayesLDM follows a Python-like syntax for the specification of constants and variables and supports standard Python syntax for mathematical operations. BayesLDM also supports the use of numerical functions from the JAX library. BayesLDM expressions are compositions of variables and constants defined using arithmetic operations and numerical functions.

## 3.2 Assignment and Distribution Statements

An assignment statement assigns a numerical value to a variable. The left hand side of the assignment statement must be a valid BayesLDM variable name. The right hand side must be a valid BayesLDM numerical expression.

A distribution statement specifies a probability distribution for a variable. The left hand side of the statement must be a valid BayesLDM variable name. The right hand side must be a valid BayesLDM probability distribution expression. A probability distribution expression consists of a distribution name and BayesLDM expressions for each of the distribution's parameters.

Example1 below specifies a basic BayesLDM model consisting of one assignment statement `b=1` and two distribution statements `s ~ Exp(b)` and `x ~ N(0,4*s)`. The assignment statement simply assigns the value 1 to the variable $b$. The first distribution statement specifies an exponential distribution on variable $s$ with parameter $b$. The second distribution statement specifies a normal distribution for $x$ with mean 0 and standard deviation $4s$. This basic model implements a scale mixture of normal distributions. BayesLDM supports a wide array of probability distributions including Bernoulli, beta, gamma, Poisson, StudentT and more.

```
ProgramName: Example1
b = 1
s ~ Exp(b)
x ~ N(0,4*s)
```

## 3.3 Arrays and Indexing

BayesLDM supports the declaration of array variables via an index-centric approach that matches the multi-level index semantics of longitudinal data. Variables used as array indices are declared via an index statement in the header of the BayesLDM model specification. Example2 below shows the basic construction of an array variable $x$ via the specification of an index variable $t$. Indexing $x$ with $t$ on the left hand side of the distribution statement implicitly declares $x$ as an array variable.

```
ProgramName: Example2
Indices: t 0 4
x[t] ~ N(0,1)
```

Any number of variables can share the same index variable, and a model can use multiple index variables. In Example3, we declare a two-level indexing structure as would match the definition of a data set consisting of multiple data cases $n$ where each data case $\mathbf{x}_n$ is a time series. In this model, each data case has its own scale parameter $s[n]$ sampled from an exponential distribution. The value at time point $t$ for data case $n$ is then sampled from a normal distribution with that data case's scale parameter $s[n]$.

```
ProgramName: Example3
Indices: n 0 4, t 0 9
s[n]   ~ Exp(1)
x[n,t] ~ N(0,s[n])
```

Importantly, indexing can also be used on the right hand side of distribution statements, which is the primary construction used to create dynamic models. In the example below, we use an index variable $t$ and let the mean of $x[t]$ depend on $x[t-1]$, creating a first-order conditionally normal autoregressive process (e.g., AR(1) model). The parameters of the model are the autoregression coefficient $a$ and the standard deviation parameter $s$. When specifying such a process, an unconditional distribution over the first time point must also be specified, thus the inclusion of the explicit distribution statement for $x[0]$.

```
ProgramName: Example4
Indices: t 0 4
a    ~ N(0,10)
s    ~ Exp(1)
x[0] ~ N(0,s)
x[t] ~ N(a*x[t-1],s)
```

The DBN model shown in Figure 3 illustrates a more complex vector autoregressive process on two variables. Below we show a BayesLDM model specification for this DBN.

```
ProgramName: Example5
Indices: t 0 19
wzz  ~ N(0,10)
wyz  ~ N(0,10)
```

```
wzy   ~ N(0,10)
wyy   ~ N(0,10)
bz    ~ N(0,10)
by    ~ N(0,10)
sz    ~ Exp(1)
sy    ~ Exp(1)
z[0]  ~ N(0,10)
y[0]  ~ N(0,10)
z[t]  ~ N(wzz*x[t-1]+wyz*y[t-1]+bz,sz)
y[t]  ~ N(wxy*x[t-1]+wyy*y[t-1]+by,sy)
```

## 4 BAYESLDM COMPILER

The model compilation process in the BayesLDM system follows a two-stage parsing and compiling process. The BayesLDM grammar is specified using the textX Python package [3], a metalanguage for specifying domain-specific languages. textX is used to automatically generate a parser for the BayesLDM language from the specification of the grammar. The parser is then applied to generate an abstract syntax tree (AST) from the model specification. In this section, we describe details of the model compilation process including compilation optimizations for specific model structures as well as how missingness is dealt with when conditioning the model on incomplete data.

### 4.1 Basic Compilation

In the case of general Bayesian network models with no indexed variables, the BayesLDM compiler iterates through the AST representation of the model and constructs an explicit directed graph representation of the model. Each node in the graph corresponds to a single random variable in the model. The distribution of each random variable, including how it depends on the values of its parents, is extracted from the AST, translated into NumPyro sampling statements using a set of templates, and stored in the corresponding graph node. The extracted graph is then processed to determine a topological order over the random variables and the sampling statements stored in the graph nodes are output in this order to produce a valid NumPyro specification of the model.

### 4.2 Optimization of Indexed Expressions

As noted previously, the BayesLDM language includes specific features for enabling compact expression of Bayesian Network and DBN models based on indexing statements and array expressions. These expressions require additional processing during model compilation. For instance, as shown in Example2, BayesLDM can use index variables and array expressions to specify implicit replication of distribution statements with respect to an index variable. The interpretation of the distribution statement `x[t] ~ N(0,1)` with $t$ ranging from 0 to 2, for example, is the following set of expressions:

```
x[0]  ~ N(0,1)
x[1]  ~ N(0,1)
x[2]  ~ N(0,1)
```

When compiling to Python and NumPyro, there are multiple options for expressing this set of operations. The most obvious is to create one node in the model graph for each index value of the variable. However, this is wasteful in terms of the length of the resulting code. We can instead compile such BayesLDM language statements using a single NumPyro *plate* statement, which provides efficient sampling of independent and identically distributed (IID) random variables.

BayesLDM also supports implicit recurrence statements of the form `x[0] ~ N(0,s)`, `x[t] ~ N(a*x[t-1],s)`. The interpretation of these distribution statements for $t$ ranging from 0 to 2, for example, is the following:

```
x[0]  ~ N(0,s)
x[1]  ~ N(a*x[0],s)
x[2]  ~ N(a*x[1],s)
```

Again, the straightforward approach is to completely expand such statements and insert them into the model graph during the compilation process. However, as for the implicit replication case noted above, this can be very inefficient. In the case where the distribution of a variable depends only on lagged copies of itself and other non-indexed variables, we leverage a NumPyro primitive referred to as *scan*. The scan primitive requires specifying a base case and a transition function as arguments and is effectively a purely functional replacement for a for loop. Leveraging the scan primitive provides a much more compact expression of the model.

### 4.3 Optimization of DBNs

When a BayesLDM model meets the definition of a Dynamic Bayesian Network, more significant optimizations are possible. In this case, the model consists of a set of non-indexed variables corresponding to the DBN parameters and a set of indexed variables corresponding to the multivariate time series data. Since the indexed variables only have dependencies backward in time and within the same time slice, the portion of the model that corresponds to the multivariate time series can be compiled into a single NumPyro scan statement where the transition function specifies how to sample all of these variables. This yields highly compact NumPyro code.

Importantly, NumPyro itself further compiles both plate and scan statements into Accelerated Linear Algebra (XLA) implementations that can provide significant speedups relative to more basic implementations of the same model structures.

### 4.4 Conditioning Models on Data

When compiling a BayesLDM model, if no data are supplied, the resulting probabilistic program implements sampling from the joint distribution over the model parameters and the data distribution. This functionality can be used for simulating data sets under the prior distribution $p(\theta)$ over the model parameters, or just for sampling sets of model parameter values from the prior.

However, BayesLDM's primary use case is inference for the posterior distribution over model parameters and missing data values when conditioning on partially observed data. To condition on data, the user supplies one or more data frames to the BayesLDM system at compile time. BayesLDM inspects the dataframes as part of the compilation process and outputs inference code that conditions on the supplied data. The Python commands for compiling a model and running sampling are shown below.

```
model = BayesLDM.compile(model_text, obs=[var1,...],
                        data=[df1,...])
samples = model.sample()
```

BayesLDM uses the Pandas library to provide a dataframe abstraction. BayesLDM inspects supplied data frames to identify column names that match variable names in the BayesLDM model. The variables to condition on are specified in the `obs` keyword argument to the compiler and the `data` keyword argument specifies the data frames. It is required that the dataframe index structure for a column that matches the name of a conditioned variable also match the indexing structure for that BayesLDM variable. Thus, if the BayesLDM model contains a conditioned variable $x[n, t]$, it is required that the data frame containing a column with name $x$ have a two-level multi-index structure where the outer level index is named $n$ and the inner level index is named $t$. Under these conditions, BayesLDM obtains index ranges for each index variable from the input data frame.

In NumPyro, an observed value for a variable can be conditioned on by passing the value to the `obs` keyword argument of the corresponding sampling statement in the NumPyro program. When BayesLDM is given a dataframe as input during compilation, it automatically sets index value ranges as noted and then inserts observed values for variables by setting the appropriate `obs` keyword values as the NumPyro program text is being generated. This makes it very easy to manipulate what data are conditioned on. When the same model has variables with different index structures, the data for each variable can be placed in separate dataframes, one with each index structure, and BayesLDM will automatically discover which dataframes contain which variables.

### 4.5 Missing Data and Bayesian Imputation

The Pandas library represents missing data in dataframes using floating point NaN values and BayesLDM follows the same convention. During the model compilation process, BayesLDM will only condition on a value contained in a dataframe if that value is not NaN. The presence of missing data interacts with more efficient replication structures in NumPyro in somewhat complicated ways. The BayesLDM compiler automatically inserts statements for conditioning only observed index values when applying compile-time optimizations.

Importantly, in the case where a variable is partially observed, the values of the variable that are not observed are automatically treated as unknown random variables by NumPyro. When the compiled BayesLDM program is executed, the observed values are conditioned on and the joint posterior distribution over all unknown parameter values and continuous missing data values is automatically computed using MCMC methods. This process implements Bayesian imputation for missing continuous variable values under the missing at random assumption as described earlier. Missing discrete random variables are currently dealt with using NumPyro's variable elimination methods.

### 4.6 Model Compilation Examples

In Appendix A, we show an example of a BayesLDM program implementing an AR(1) model along with versions compiled with and without optimization. Without optimization, the autoregressive chain is completely unrolled. With optimization, the model is expressed using the NumPyro Scan primitive. Both implementations account for missing data with the unoptomized version using Python if statements to check for missing data values and the Scan version using more complex structures to meet NumPyro's functional programming restrictions when implementing Scan transition functions.

## 5 EVALUATING MODEL REPRESENTATIONS

We now turn to the evaluation of different facets of BayesLDM. In this section we evaluate the representational efficiency of the BayesLDM modeling language using a set of benchmark models.

**Evaluation Protocol:** We select several models from the text

*Table 1.* Representational Efficiency (lines/characters of code)

| MODEL | STAN | PYMC3 | NUMPYRO | BAYESLDM |
|---|---|---|---|---|
| LINEAR REG. | 18/270 | 6/220 | 6/252 | 6/94 |
| BIN. LOGITS | 21/295 | 4/173 | 4/183 | 5/105 |
| MULTI A | 24/370 | 6/261 | 6/303 | 7/143 |
| MULTI B | 32/679 | 10/516 | 10/644 | 12/328 |
| ZERO-INF. | 22/414 | 6/228 | 6/251 | 5/106 |
| AR(2) | 23/369 | 6/237 | 12/505 | 9/160 |
| AVERAGE | 23.3/399.5 | 6.3/272.5 | 7.3/356.3 | 7.3/156.0 |

*Statistical Rethinking* as a modeling benchmark as these models have existing reference implementations in each of the probabilistic programming languages that we compare to [14]. The models that we consider are Linear Regression, Binomial Logits, Multi-Level A, Multi-Level B, and Zero-Inflated [14] p.131, 417, 423, 429 and 390. We include AR(2) as a longitudinal model.

We compare BayesLDM to three probabilistic programming languages: NumPyro [19], Stan [2], and PyMC3 [20]. We use reference implementations for the *Statistical Rethinking* models provided for NumPyro[1], PyMC3[2], and Stan[3]. For the AR(2) model, we use reference implementations from the documentation of NumPyro[4], PyMC3[5] (built-in $AR$ function), and Stan[6]. The BayesLDM implementations of these models are provided in Appendix A. We note that we made minor modifications to the variable names in the reference implementations of all models so that all implementations use the same variable names. Source code for all models is available at https://github.com/reml-lab/BayesLDM. To assess representational efficiency, we consider both the number of lines of code needed to specify a model as well as the number of characters of code.

**Results:** The results of this assessment are shown in Table 1. As we can see, the Stan implementations require two to three times the number of lines of code to specify the same model compared to PyMC3, NumPyro and BayesLDM. This is largely due to the fact that Stan requires declaring variables before use and partitions model and data specification across different program sections. In terms of character count, we can see that BayesLDM model representations are approximately 50% more compact than Stan and NumPyro and approximately 40% more compact that PyMC3. These results show that BayesLDM can be significantly more representationally efficient when specifying the same model. BayesLDM thus has the potential to make models easier to specify, read and update.

## 6 EVALUATING INFERENCE SCALABILITY

In this section we evaluate the scalability of inference code produced by the BayesLDM model compiler.

### 6.1 Experiment 1: Benchmark Models

**Experimental Protocol:** For this experiment, we use the set of models introduced in the previous section and their reference implementations in Stan, PyMC3 and NumPyro along with the BayesLDM implementations of these models. We use the data sets from [14]. For the AR(2) model, we use the data set introduced in the NumPyro documentation of this model. For all models, we evaluate the inference run time of the NUTS sampler using 1000 sampling iterations following 500 burn-in iterations. Experiments were run on Google Colab with the Google Compute Engine backend.

---

[1] https://fehiepsi.github.io/rethinking-numpyro/
[2] https://github.com/pymc-devs/resources/tree/master/Rethinking_2
[3] https://vincentarelbundock.github.io/rethinking2/
[4] https://num.pyro.ai/en/stable/examples/ar2.html
[5] https://docs.pymc.io/en/v3/pymc-examples/examples/time_series/AR.html
[6] https://mc-stan.org/docs/2_28/stan-users-guide/autoregressive.html/

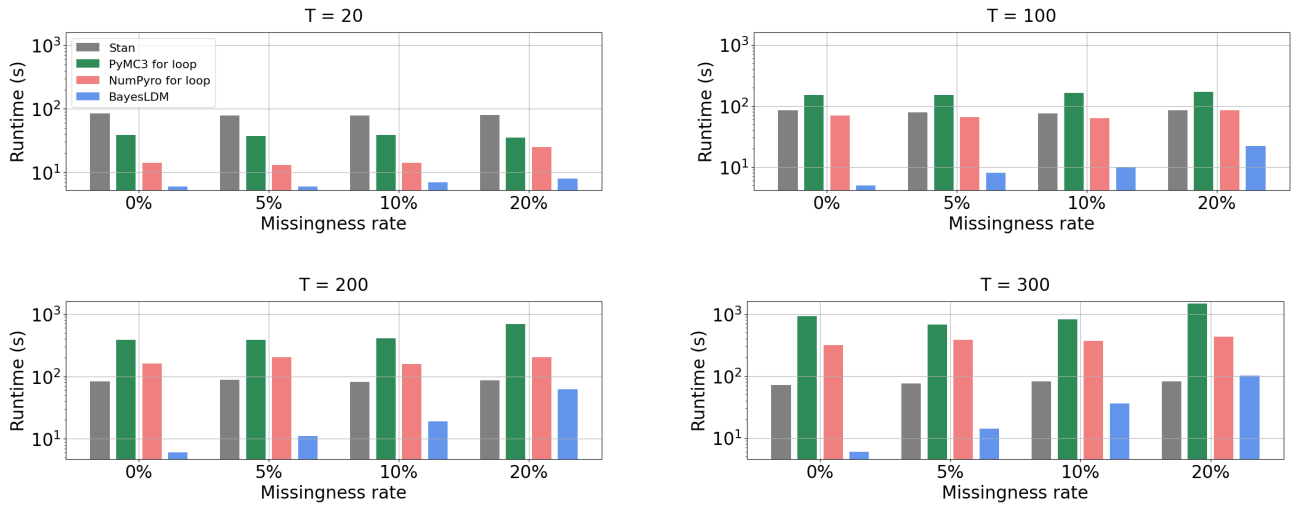

*Figure 4.* AR(1) inference scalability using Stan, PyMC3, NumPyro (for loop), and BayesLDM

*Table 2.* Inference Scalability (seconds)

| MODEL | STAN | PYMC3 | NUMPYRO | BAYESLDM |
|---|---|---|---|---|
| LINEAR REGRESSION | 67.4 | 70.3 | 8.9 | 9.4 |
| BINOMIAL LOGITS | 72.1 | 80.4 | 8.5 | 9.6 |
| MULTI-LEVEL A | 73.2 | 42.3 | 9.1 | 10.3 |
| MULTI-LEVEL B | 89.7 | 132.7 | 12.9 | 13.4 |
| ZERO-INFLATED | 83.4 | 114.8 | 12.8 | 13.1 |
| AR(2) | 75.4 | 83.9 | 11.1 | 11.5 |
| AVERAGE | 76.9 | 87.4 | 10.8 | 11.2 |

*Table 3.* AR(1) Model Representation Statistics

| MODEL | STAN | PYMC3 | NUMPYRO | BAYESLDM |
|---|---|---|---|---|
| AR(1)+MISS | 29/531 | 16/572 | 19/750 | 7/125 |

**Results:** The results of this experiment are shown in Table 2. As we can see, the BayesLDM and corresponding reference NumPyro implementation are generally within one second of each other. We note that for the AR(2) model, the NumPyro reference implementation uses scan to efficiently iterate over time points resulting in fast execution time. However, the BayesLDM compiler outputs optimized inference code that also uses scan and thus matches the run time obtained by the NumPyro reference implementation while permitting a model specification that is much more compact. Next, we observe that PyMC3 and Stan are 7 to 8 times slower than NumPyro and BayesLDM on this benchmark. This appears to be partly attributable to both of these frameworks taking a substantial amount of time to compile models. However, as we will see in the next section, Stan's run time is stable for larger data sets, while PyMC3's becomes worse.

### 6.2 Experiment 2: AR(1) with Missing Data

**Experimental Protocol:** As a testbed model for evaluating missing data handling capabilities of BayesLDM, we select a single-variable lag 1 autoregressive model AR(1). We generate a synthetic data set consisting of observations for 300 time points using an AR(1) model $y[t] = \mathcal{N}(a * y[t-1] + b, \sigma^2)$ with the parameter values $a = 0.9$, $b = 0.1$ and $\sigma = 0.5$. We then create missing data completely at random with several missingness rates from 5% to 20% to illustrate how inference time scales as a function of missing data rate. We also vary the length of the data available for inference from the first 20 time points to all 300 time points. Note that increasing data length does not change the dimension of the underlying Markov chain state space, while increasing the missing data rate does increase the dimension of the underlying Markov chain state space. In all cases, we run the NUTS sampler for 1000 sampling iterations following 500 burn-in iterations. We assess the total inference run time in seconds.

**Model Implementations:** We compare implementations of the AR(1) model with missing data in Stan, PyMC3, NumPyro and BayesLDM. For NumPyro, we compare to a basic implementation of inference using a for loop over time points as might be produced by a non-expert user. The run time of a NumPyro implementation using scan is again identical to that obtained by BayesLDM and is not shown. All AR(1) model implementations are shown in Appendix B. Note that BayesLDM natively supports marginalization over missing data during inference and thus the model specification is the same regardless of whether data are missing or not. We briefly summarize the representation statistics for these implementations in the table below. As we can see, BayesLDM requires two to four times fewer lines and four to six times fewer characters of code.

**Results:** Figure 4 compares the inference run times for the four implementations (note that we use a log y-axis in these plot). For $T = 20$, the NumPyro for loop and BayesLDM implementations are both more efficient than Stan and PyMC3, likely due to the compilation overhead issues noted previously. However, as the data set size grows, the inefficiency in the NumPyro for loop implementation quickly becomes apparent and the optimized BayesLDM scan implementation can be many times faster. We see that at $T = 200$ and $T = 300$, the NumPyro for loop implementation becomes much slower than Stan, while PyMC3 becomes slower than Stan at T=100 and its performance degrades substantially for longer sequence lengths. Finally, we can see that the compilation process in Stan does pay off eventually as it slightly outperforms BayesLDM for T=300 at the largest missing data rate considered.

These results clearly show that the compile time optimizations performed by BayesLDM lead to dramatically faster inference times compared to the basic for loop implementation that a non-expert NumPyro user is likely to produce when implementing such models. We can also see potential for a meta-optimization strategy for BayesLDM where depending on characteristics of the model and data set, BayesLDM could choose to compile into implementations for different probabilistic programming languages. We leave this as a direction for future work. Finally, we note that all four implementations do produce equivalent inference results, as expected. In Figure 6 in Appendix C, we show sample posterior plots produced using all implementations for $T = 300$ time points and 20% missingness rate for reference.

## 7 MODELING CASE STUDY

In this section, we present a modeling case study using real data from a physical activity intervention trial in the mobile health application domain to further illustrate the use of BayesLDM. The primary goal of the trial was to collect data to begin to understand the relationship between physical activity and a variety of associated behavioral constructs. One of the intervention components used in this study focuses on physical activity planning. Participants are asked to plan physical activity or exercise sessions at the beginning of each week. We apply BayesLDM to model these data using a hierarchy of increasingly complex model structures.

**Variables:** The specific planning variable of interest in this study is the number of minutes of planned physical activity per week ($P$). A Fitbit device is used to assess the actual number of minutes of physical activity per week ($A$). The behavioral variables that are hypothesized to be relevant are competence for physical activity ($C$), extrinsic motivation ($EM$), and intrinsic motivation ($IM$).

**Data Set:** This IRB-approved study took place over 38 weeks with weekly observations of the three behavioral constructs and two activity-related variables, among other data. The planning activity primarily occurred on Sunday evenings and self-report questionnaires were also answered at that time. Here we analyze data from 10 study participants to illustrate the use of BayesLDM. In Table 4 we show a summary of the missingness rates for each of these variables, across the participants. We note that if a participant does not create an activity plan, the number of planned minutes is 0 and thus this variable is always observed. We can see that the average missingness rate for the behavioral variables is 16 to 20%, while the maximum rates over all participants is much higher. The number of actual minutes of activity in a given week is passively sensed and thus the missingness rate is lower. However, there are instances of participants not wearing the Fitbit device for an extended time, leading to some missingness. Lastly, we note that the minimum missing rate for all variables is 0%.

**Hypotheses:** The initial set of hypotheses in this domain were as follows: (1) each of the variables in this domain has an autoregressive dependence where the values at each time step are influenced by the previous time step; (2) extrinsic motivation and intrinsic motivation for physical activity positively influence the number of physical activity minutes that a participant plans in a given week; (3) the number of physical activity minutes that are planned in a given week positively influences the number of minutes of physical activity that are actually completed in a given week; (4) the number of actual minutes of physical activity completed positively influences competency for physical activity, and (5) competency for physical activity positively influences intrinsic motivation. Since planning occurs at the start of a week, it is hypothesized to directly influence activity levels within the same week. All other hypothesized relationships express influence across weeks. Finally, due to heterogeneity among study participants, it is hypothesized that (6) the degree of influence between variables may vary from person to person.

*Table 4.* Summary of missingness rates across 10 participants.

| VARIABLE | MAX | AVERAGE |
|---|---|---|
| PLANNED MINUTES ($P$) | 0% | 0% |
| ACTUAL MINUTES ($A$) | 10% | 2% |
| COMPETENCY ($C$) | 50% | 20% |
| EXTRINSIC MOTIVATION ($EM$) | 52% | 16% |
| INTRINSIC MOTIVATION ($IM$) | 52% | 16% |

**Models:** We operationalize the hypotheses described above into a set of increasingly detailed models. We begin with a model that expresses only hypothesis (1): at each time point, each variable depends on its own previous value. This multivariate AR(1) model is shown in Figure 5 (left). The next model that we posit adds the additional dependencies between variables expressed in hypotheses (2) to (5), converting the AR(1) model to a general DBN. This model is shown in Figure 5 (right). The final model that we posit allows the parameters on the relationship between planning and actual activity to be participant-specific. This is a crucial link in the model that relates the level of planning to actual activity and is likely to vary from person to person, reflecting hypothesis (6). We apply a common prior to the participant-specific parameters resulting in a multi-level structure for these parameters. We refer to this model as DBN-ML. To complete the specification of these models, we need to specify distributions for each variable. We select linear Gaussian components for all relationships between data variables. The weight and bias model parameters are given normal priors, while the standard deviation parameters have exponential priors. The BayesLDM implementation for the DBN is shown in Appendix D. Each time slice represents one study week.

**Evaluation Metrics:** We test the goodness of fit of each model using the negative log likelihood (NLL), AIC and BIC scores. It is well known that computing the log likelihood on the data used to train a model will over estimate performance for models with more parameters. The BIC and AIC scores attempt to correct for this by penalizing models based on their parameter count and the number of observations. Lower values of all three scores indicate better model fits.

**Results:** We show the results for the AR(1), DBN, and DBN-ML models in Table 5. As we can see, the DBN model provides substantially better fit to the data according to all three goodness of fit measures. Adding the multi-level structure to the parameters of the planning-to-activity link results in an improvement in NLL and AIC, but an increase in BIC due to the more severe penalization for adding additional parameters using BIC. These results indicate that the DBN is a significant improvement over the AR(1) model, while the added complexity of the multi-level extension may not be justified given the volume of data.

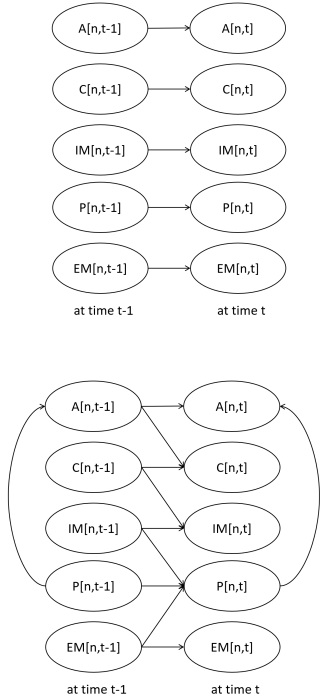

*Figure 5.* Specification of the AR(1) model (top) and the DBN (bottom) model for activity planning.

*Table 5.* Goodness of fit statistics for the planning domain.

| MODEL | NLL | BIC | AIC | RUNTIME (S) |
|---|---|---|---|---|
| AR(1) | 2387.6 | 4886.7 | 4805.3 | 423.1 |
| DBN | 2250.7 | 4649.9 | 4541.3 | 459.0 |
| DBN-ML | 2225.9 | 4682.1 | 4513.8 | 484.9 |
| DBN-SIMPLIFIED | 2247.9 | 4592.4 | 4521.8 | 433.0 |
| DBN-SIMPLIFIED-ML | 2222.9 | 4624.0 | 4493.7 | 473.7 |

Figures 7 and 8 in Appendix D show the posterior samples and MCMC summary for select model parameters in the DBN model. The parameters in the model that correspond to the primary hypothesized relationships are $w\_ep$, $w\_ip$, $w\_pa$, $w\_ac$, and $w\_ci$. As we can see, the posterior distributions for the parameters $w\_ep$, $w\_pa$, and $w\_ci$ have almost all of their mass on positive values. On the other hand, $w\_ac$ and $w\_ip$ have posterior distributions concentrated near zero. This analysis provides evidence that for these data, the hypothesized relationships between actual activity and competence ($w\_ac$), and between intrinsic motivation and planning ($w\_ip$) may not hold. Lastly, we can also see that all of the bias parameters have distributions that concentrate around zero.

We can attempt to remove the parameters with posteriors concentrated around zero from the DBN and DBN-ML models and re-assess the goodness of fit. The resulting models are DBN-Simplified and DBN-Simplified-ML. As we can see in Table 5, DBN-Simplified results in improved goodness of fit for all metrics relative to the initial DBN model. This result provides further evidence that the pruned relationships may not hold in this data set. Again, we can see that DBN-Simplified-ML improves on DBN-Simplified in terms of NLL and AIC, but not in terms of the more stringent BIC criteria indicating only weak support for the hypothesis that different participants are likely to express different relationships between planning and activity.

Lastly, we note that Table 5 also shows inference run times for all models. As we can see, increasing the complexity of the models increases their inference run time, but inference executes for all models in less than 10 minutes. As next steps in such an analysis a user might look at incorporating additional observed variables, inspect residuals for individual variables to assess the need for non-linearities in the model, or extend the model to incorporate hypothesized missing data mechanisms [10]. All of these tasks can be accomplished using BayesLDM.

## 8 CONCLUSION

In this paper we have presented BayesLDM, a system for longitudinal data modeling consisting of a high-level modeling language with specific features for modeling complex multivariate time series data coupled with a compiler that can produce optimized code for performing probabilistic inference in the specified model. We have demonstrated that the BayesLDM modeling language can be substantially more representationally efficient than existing PPLs when while simultaneously producing optimized inference code that is highly efficient.

Next, we have demonstrated the ability of BayesLDM to automatically deal with missing data, greatly simplifying the task of producing performant inference code for time series models that must be applied to incomplete data. Finally, we have illustrated the application of BayesLDM to a real-world modeling task subject to data scarcity and missingness, showing how it can be used to encode domain hypotheses and support iterative model refinement.

In summary, we believe that the capabilities of BayesLDM have the potential to significantly accelerate modeling workflows by abstracting away the process of producing computationally efficient probabilistic inference code for analyzing complex real-world time series data. We hope that BayesLDM will be a useful modeling tool across a wide range of application domains including the modeling and analysis of complex longitudinal health data sets.

## ACKNOWLEDGEMENTS

This work is supported by National Institutes of Health National Cancer Institute, Office of Behavior and Social Sciences, and National Institute of Biomedical Imaging and Bioengineering through grants U01CA229445 and 1P41EB028242.

## A  BENCHMARK MODELS IN BAYESLDM

In this section, we provide BayesLDM implementations of the benchmark models used in the study of representational efficiency.

```
ProgramName: LinearRegression
Inputs: x
a     ~ N(0,.2)
bM    ~ N(0,.5)
sigma ~ Exp(1)
y     ~ N(a + bM * x, sigma)
```

```
ProgramName: BinomialLogits
Indices: j 0 47, i 0 47
Inputs: tank, D
a[j] ~ N(0,1.5)
S[i] ~ BinomialLogits(D[i], a[tank[i]])
```

```
ProgramName: MultiLevelA
Indices: j 0 47, i 0 47
Inputs: tank, D
mu    ~ N(0,1.5)
sigma ~ Exp(1)
a[j] ~ N(mu,sigma)
S[i] ~ BinomialLogits(D[i], a[tank[i]])
```

```
ProgramName: MultiLevelB
Indices: j 0 6, k 0 5, l 0 3, i 0 503
Inputs: actor, block_id, treatment
a_bar   ~ N(0,1.5)
sigma_a ~ Exp(1)
sigma_g ~ Exp(1)
sigma_b ~ Exp(1)
a[j] ~ N(a_bar,sigma_a)
g[k] ~ N(0,sigma_g)
b[l] ~ N(0,sigma_b)
logit_p[i] = a[actor[i]] + g[block_id[i]] + b[treatment[i]]
pulled_left[i] ~ BinomialLogits(1,logit_p[i])
```

```
ProgramName: ZeroInflated
Indices: n 0 199
ap ~ N(-1.5,1)
al ~ N(1,0.5)
y  ~ ZeroInflatedPoisson(expit(ap),exp(al))
```

```
ProgramName: AR2
Indices: t 0 141
a0 ~ N(0,1)
a1 ~ N(0,1)
a2 ~ N(0,1)
s  ~ HalfNormal(1)
y[0] ~ N(0,1)
y[1] ~ N(0,1)
y[t] ~ N(a0 + a1 * y[t-1] + a2 * y[t-2], s)
```

## B  IMPLEMENTATIONS OF THE AR(1) MODEL

Below we provide implementations of the AR(1) model, including missing data handling, used in inference run time experiments.

```
BayesLDM AR1 Implementation:

ProgramName: AR1
Indices: t 0 299
a     ~ N(0,10)
b     ~ N(0,10)
sigma ~ HalfNormal(10)
y[0] ~ N(0,10)
y[t] ~ N(a*y[t-1] + b, sigma)
```

```
Compiled BayesLDM AR1 Implementation (with no optimization):

def AR1(df=None):
  a=numpyro.sample("a",dist.Normal(0,10))
  b=numpyro.sample("b",dist.Normal(0,10))
  sigma=numpyro.sample("sigma",dist.HalfNormal(10.0))
  y_0=numpyro.sample("y[0]",dist.Normal(0,10),
    obs=None if (np.isnan(df["y"].loc[0])) else df["y"].loc[0])
  y_1=numpyro.sample("y[1]",dist.Normal(a*y_0+b,sigma),
    obs=None if (np.isnan(df["y"].loc[1])) else df["y"].loc[1])
```

```
.
.
y_299=numpyro.sample("y[299]",dist.Normal(a*y_298+b,sigma),
    obs=None if (np.isnan(df["y"].loc[299])) else df["y"].loc[299])
```

```
Compiled BayesLDM AR1 Implementation (with optimization):

def AR1(df=None):
 a=numpyro.sample("a",dist.Normal(0,10))
 b=numpyro.sample("b",dist.Normal(0,10))
 sigma=numpyro.sample("sigma",dist.HalfNormal(10))
 y_data=np.array(df["y"].values)
 y_0=numpyro.sample("y[0]",dist.Normal(0,10),
    obs=y_data[0] if (not np.isnan(y_data[0])) else None)
 y_impute=jnp.zeros((len(y_data)))
 y_nan_indices=[n for n in np.argwhere(
                np.isnan(y_data)).flatten() if n >= 1]
 for i in y_nan_indices:
    y_impute=y_impute.at[i].set(numpyro.sample("y"+str([i]),
            dist.Normal(0,10).mask(False)))

 def transition(carry,value):
    y_tm1=carry[0]
    t_carry=carry[1]
    y_t_obs=jnp.where(jnp.any(t_carry==jnp.array(y_nan_indices)),
            y_impute[t_carry],value)
    y_t=numpyro.sample("y_t",dist.Normal(a*y_tm1+b,sigma),
        obs=y_t_obs)
    return (y_t,t_carry+1),(y_t)

 carry_start=(y_0,1)
 _,(y)= scan(transition,carry_start,y_data[1:])
```

```
Stan AR1 Implementation:

data {
    int<lower=0> To;
    int<lower=0> Tm;
    int<lower=1,upper=To+Tm> ii_o[To];
    int<lower=1,upper=To+Tm> ii_m[Tm];
    real y_obs[To];
}
transformed data {
    int<lower=0> T=To+Tm;
}
parameters {
    real a;
    real b;
    real<lower=0>sigma;
    real y_mis[Tm];
}
transformed parameters {
    real y[T];
    y[ii_o] = y_obs;
    y[ii_m] = y_mis;
}
model {
    a ~ normal(0,10);
    b ~ normal(0,10);
    sigma ~ normal(0,10);
    y[1] ~ normal(0,10);
    for (t in 2:T)
        y[t] ~ normal(a*y[t-1]+b,sigma);
}
```

```
NumPyro AR1 For Loop Implementation:

def AR1(df=None):
    a = numpyro.sample('a',dist.Normal(0,10))
    b = numpyro.sample('b',dist.Normal(0,10))
    sigma=numpyro.sample('sigma',dist.HalfNormal(10))
    y = jnp.zeros((300))
    value=df['y'].loc[df.index.unique('t')[0]]
    if str(value).lower() == 'nan':
        y = y.at[0].set(numpyro.sample('y'+str([0]),
            dist.Normal(0,10)))
    else:
        y = y.at[0].set(numpyro.sample('y'+str([0]),
            dist.Normal(0,10), obs=value))
    for t,t_ in enumerate(df.index.unique('t')[1:],1):
        value=df['y'].loc[t_]
        if str(value).lower() == 'nan':
            y = y.at[t].set(numpyro.sample('y'+str([t]),
                dist.Normal(a*y[t-1]+b,sigma)))
        else:
            y = y.at[t].set(numpyro.sample('y'+str([t]),
                dist.Normal(a*y[t-1]+b,sigma),obs=value))
```

```
PyMC3 AR1 Implementation:

with pm.Model() as AR1:
    a = pm.Normal('a',0.0,0.0)
    b = pm.Normal('b',0.0,10.0)
    sigma = pm.HalfNormal("sigma",10.)
    y_data = np.array(df["y"].values)
    if str(y_data[0]).lower() != 'nan':
        y = pm.Normal('y'+str([0]),0.,10., observed=y_data[0])
    else:
        y = pm.Normal('y'+str([0]),0.,10.)
    for t in range(1, len(y_data)):
        mu = b + pm.math.dot(y, a)
        if str(y_data[t]).lower() != 'nan':
            y = pm.Normal('y'+str([t]),mu,sigma,
                                    observed=y_data[t])
        else:
            y = pm.Normal('y'+str([t]),mu,sigma)
```

## C  INFERENCE SCALABILITY MATERIALS

In Figure 6 we show additional visualizations confirming that all model implementations provide the same results.

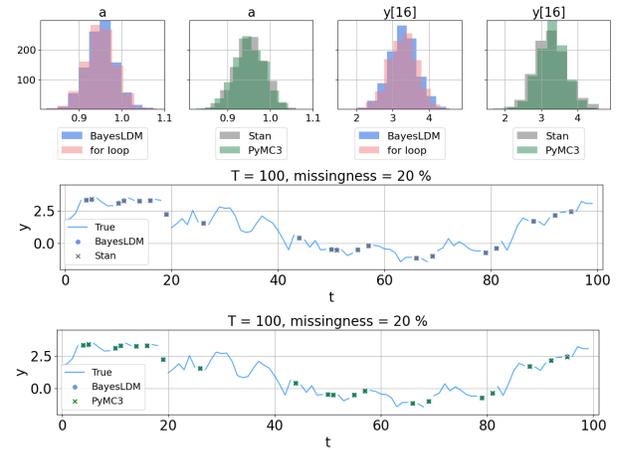

*Figure 6.* Selected posterior marginals (top). True data and imputed data (middle, bottom).

## D  CASE STUDY MATERIALS

This section includes the supporting materials for the planning domain modeling case study. Below we show the initial planning domain DBN model implementation in BayesLDM. We show the MCMC summary for this model in Figure 8 and posterior samples for select model parameters in Figure 7.

```
ProgramName: DBN
Indices: n 0 9, t 0 37
w_ee  ~ N(0,10)
b_e   ~ N(0,10)
w_ci  ~ N(0,10)
w_ii  ~ N(0,10)
b_i   ~ N(0,10)
w_pp  ~ N(0,10)
w_ep  ~ N(0,10)
w_ip  ~ N(0,10)
b_p   ~ N(0,10)
w_pa  ~ N(0,10)
w_aa  ~ N(0,10)
b_a   ~ N(0,10)
w_ac  ~ N(0,10)
w_cc  ~ N(0,10)
b_c   ~ N(0,10)
s_e   ~ Exp(1)
s_i   ~ Exp(1)
s_a   ~ Exp(1)
```

```
s_c      ~ Exp(1)
s_p      ~ Exp(1)
EM[n,0]  ~ N(0,10)
IM[n,0]  ~ N(0,10)
A[n,0]   ~ N(0,10)
C[n,0]   ~ N(0,10)
P[n,0]   ~ N(0,10)
EM[n,t]  ~ N(w_ee * EM[n,t-1] + b_e, s_e)
IM[n,t]  ~ N(w_ci * C[n,t-1] + w_ii * IM[n,t-1] + b_i, s_i)
P[n,t]   ~ N(w_pp * P[n,t-1] + w_ep * EM[n,t-1] + w_ip * IM[n,t-1]
             + b_p, s_p)
A[n,t]   ~ N(w_pa * P[n,t]   + w_aa * A[n,t-1]  + b_a, s_a)
C[n,t]   ~ N(w_ac * A[n,t-1] + w_cc * C[n,t-1]  + b_c, s_c)
```

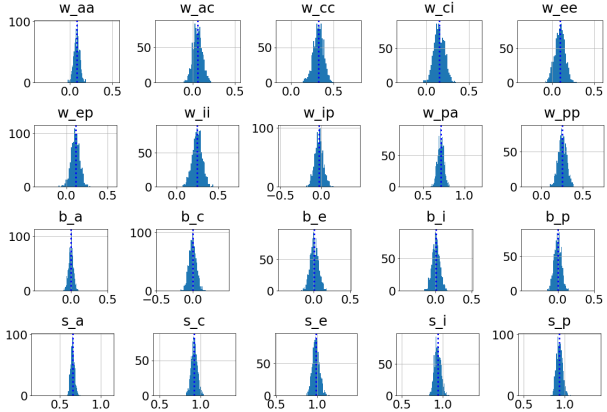

Figure 7. Histograms of selected posterior samples for the initial planning DBN model.

|      | mean  | std  | median | 5.0%  | 95.0% | n_eff   | r_hat |
|------|-------|------|--------|-------|-------|---------|-------|
| b_a  | 0.00  | 0.03 | 0.00   | -0.06 | 0.06  | 1966.59 | 1.00  |
| b_c  | 0.00  | 0.05 | -0.00  | -0.08 | 0.09  | 2287.96 | 1.00  |
| b_e  | 0.01  | 0.06 | 0.01   | -0.09 | 0.10  | 1535.93 | 1.00  |
| b_i  | 0.01  | 0.05 | 0.00   | -0.08 | 0.09  | 1466.66 | 1.00  |
| b_p  | 0.00  | 0.05 | 0.00   | -0.08 | 0.08  | 1821.43 | 1.00  |
| s_a  | 0.66  | 0.03 | 0.66   | 0.62  | 0.70  | 1751.56 | 1.00  |
| s_c  | 0.92  | 0.04 | 0.92   | 0.86  | 0.99  | 1321.83 | 1.00  |
| s_e  | 0.99  | 0.04 | 0.99   | 0.92  | 1.05  | 1513.35 | 1.00  |
| s_i  | 0.94  | 0.04 | 0.93   | 0.87  | 1.00  | 1483.19 | 1.00  |
| s_p  | 0.93  | 0.04 | 0.93   | 0.88  | 0.99  | 2016.98 | 1.00  |
| w_aa | 0.08  | 0.04 | 0.08   | 0.02  | 0.14  | 1895.05 | 1.00  |
| w_ac | 0.06  | 0.05 | 0.06   | -0.02 | 0.15  | 1130.53 | 1.01  |
| w_cc | 0.32  | 0.06 | 0.32   | 0.22  | 0.41  | 1362.14 | 1.00  |
| w_ci | 0.15  | 0.06 | 0.15   | 0.06  | 0.26  | 792.80  | 1.00  |
| w_ee | 0.10  | 0.06 | 0.10   | 0.00  | 0.20  | 1078.92 | 1.00  |
| w_ep | 0.11  | 0.05 | 0.11   | 0.03  | 0.20  | 1899.37 | 1.00  |
| w_ii | 0.25  | 0.06 | 0.25   | 0.15  | 0.34  | 906.59  | 1.00  |
| w_ip | -0.03 | 0.05 | -0.02  | -0.11 | 0.06  | 1464.02 | 1.00  |
| w_pa | 0.71  | 0.04 | 0.71   | 0.66  | 0.78  | 1955.73 | 1.00  |
| w_pp | 0.26  | 0.05 | 0.26   | 0.18  | 0.34  | 1812.06 | 1.00  |

Figure 8. MCMC summary for the initial planning DBN.